# ROBUST POSITIONING OF DRONES FOR LAND USE MONITORING IN STRONG TERRAIN RELIEF USING VISION-BASED NAVIGATION

* Oleg Kupervasser[1,2], Vitalii Sarychev[2], Alexander Rubinstein[2] and Roman Yavich[1]

[1] Department of Mathematics, Ariel University, Israel; [2] TRANSIST VIDEO LLC, Skolkovo, Russia

**ABSTRACT:** For land use monitoring, the main problems are robust positioning in urban canyons and strong terrain reliefs with the use of GPS system only. Indeed, satellite signal reflection and shielding in urban canyons and strong terrain relief results in problems with correct positioning. Using GNSS-RTK does not solve the problem completely because in some complex situations the whole satellite's system works incorrectly. We transform the weakness (urban canyons and strong terrain relief) to an advantage. It is a vision-based navigation using a map of the terrain relief. We investigate and demonstrate the effectiveness of this technology in Chinese region Xiaoshan. The accuracy of the vision-based navigation system corresponds to the expected for these conditions. It was concluded that the maximum position error based on vision-based navigation is 20m and the maximum angle Euler error based on vision-based navigation is 0.83 degree. In case of camera movement, the maximum position error based on vision-based navigation is 30m and the maximum Euler angle error based on vision-based navigation is 2.2 degrees.

## 1. INTRODUCTION

Recently International Journal of Geomate has published a very important paper about the integration of Geography Information System (GIS) and Global Navigation Satellite System – Real Time Kinematics (GNSS-RTK) for land use monitoring [1].
Indeed, the authors of [1] write: "limited open space area requires accurate monitoring to maintain changes in land use that is not suitable for city spatial planning. The difference between spatial planning and existing land Use should be minimized". The main problems are robust positioning in urban canyons and strong terrain reliefs with the use of GPS system only. Indeed, satellite signal reflection and shielding in urban canyons and strong terrain reliefs results in problems with correct positioning. "In urban area study, signal problems become obstacles for determining the position and navigation due to some problems such as shadowing and multipath effects" [1,2].

The paper suggests using GNSS-RTK (a full system which is a constellation of current satellites such as GPS (Global Positioning System-USA), GLONASS (Russia), Galileo (Europe), BeiDou (China), and QZSS (Japan)) that can improve positioning [1,3]. However, it does not solve the problem completely, because in some complex situations the whole satellite's system works incorrectly.

We need to transform the weakness (urban canyons and strong terrain reliefs) to an advantage. It is a vision-based navigation using a map of the terrain relief. This topic had been recently widely developed by the authors [4-10]. Vision-based algorithms have been a major research issue during the past decades. Two common approaches for the navigation problem are the landmarks [11,12]. And the ego-motion integration [13-17].

In [18], the drift is being corrected with the help of a Digital Terrain Map (DTM). The DTM is a discrete representation of the observed ground's topography. It contains the altitude over the sea level of the terrain for every geographical location. A patch from the ground was reconstructed using `structure-from-motion' (SFM) algorithm and was matched to the DTM in order to derive the camera's pose. The algorithm presented in this work [4-9] does not require an intermediate explicit reconstruction of the 3D world. By combining the DTM information directly with the images information it is claimed that the algorithm is well-conditioned and generates accurate estimates for reasonable scenarios and error sources.

Comparison of the corrected position of the object, measured from the data of Google Earth in China, with the calculated position of the object would estimate the real effectiveness of navigation corrections. The correspondent investigation for the described method was carried out during the flight in Galilee in Israel [19]. The position error was about 25 meter and angle error was about 1.5 degree.

Based on the core theory [4-10], TRANSIST VIDEO LLC (Russian Skolkovo company under the leadership of Kupervasser) developed the computer program "Vision-based navigation of UAV over relief" [20]. This program was tested in Zhejiang Province in east China near the capital Hangzhou using Google Earth data. This work was funded by Hangzhou AVISI Electronics Co. LTD. Currently, TRANSIST VIDEO LLC won the grant from the governing body of Hangzhou for the creation of the real-time version of the program. This program will be developed in collaboration with Ariel University in Israel.

## 2. METHODOLOGY

### 2.1 Finding of the corresponding characteristic points on the first and second shots (optical-flow field)

Two methods are used to choose the characteristic points on the first shot:
1) the characteristic points on the first shot are chosen on the regular grid on the boundaries of the shot. The grid is square to the size of N x N, where $N^2$ is the general number of the characteristic points.
2) the characteristic points on the first shot are chosen with the help of Shi Tomassi corner detector. The Shi Tomassi corner detector [21] (Shi-Tomasi or Kanade-Tomasi, 1993) works as follows: for the given picture let us consider the window (usually the size of the window is a 5x5 pixel, but it can depend on the size of the picture) in the center $(x_0, y_0)$. Let's determine $M$ – autocorrelation matrix:

$$M(x_0, y_0) = \sum_{(x,y) \in W} \omega(x - x_0, y - y_0) \begin{bmatrix} I_x^2(x,y) & I_x(x,y) I_y(x,y) \\ I_x(x,y) I_y(x,y) & I_y^2(x,y) \end{bmatrix}$$

(1)

where $w(x,y)$ is a weight function (usually the Gaussian function or a binary window is used) (Fig.1).

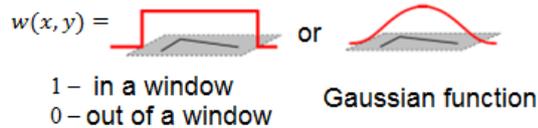

Fig 1.   Weight function

A corner is characterized by large absolute values of the matrix eigenvalues M: $M: \lambda_1 \gg 0$, $\lambda_2 \gg 0$. Shi and Tomassi suggested the measure of the corner: $F(x,y) = \min(\lambda_1, \lambda_2)$. The function finds the angles with large eigenvalues.

We use the same equation to analyze the Lucas and Kanade optical flow[22], this local method of optical flow measurement widely used in the computer vision.

The optical-flow field is supplied: $\{u_i(t_k)\}$ ($i=1...n$, $k=1,2$). For the $i$'the feature, $u_i(t_1) \in \mathbb{R}^2$ and $u_i(t_2) \in \mathbb{R}^2$ represent its locations in the first and second frame respectively:

$$\begin{bmatrix} u_i^x(t_k) \\ u_i^y(t_k) \end{bmatrix} = M^{-1}(x_i, y_i) \begin{bmatrix} -\sum_{(x,y) \in W} \omega(x - x_i, y - y_i) I_x(x,y,t_k) I_t(x,y,t_k) \\ -\sum_{(x,y) \in W} \omega(x - x_i, y - y_i) I_y(x,y,t_k) I_t(x,y,t_k) \end{bmatrix}$$

(2)

We use affine or translation monitoring of the first shot points on the second shot using pyramid realization (affine or translation) of Lucas and Kanade tracker. [22].

### 2.2 The Navigation Algorithm of DTM usage and optical flow finding

The problem can be briefly described as follows: At any given time instance $t$, a coordinate system $C(t)$ is fixed to a camera in such a way that the $Z$-axis coincides with the optical axis and the origin coincides with the camera's projection center. At that time instance $t_1$ the camera is located at some geographical location $p_1 = p(t_1)$ and has a given orientation $R_1 = R(t_1)$ with respect to a global coordinate system $W$ ($p(t)$ is a 3D vector, $R(t)$ is an orthonormal rotation matrix). Consider now two sequential time instances $t_1$ and $t_2$: the transformation from $C(t_1)$ to $C(t_2)$ is given by the translation vector $p_{12} = \Delta p(t_1, t_2)$ and the rotation matrix $R_{12} = \Delta R(t_1, t_2)$. Let $q_{1i} = {}^{C_i}q(t_1)$ and $q_{2i} = {}^{C_i}q(t_2)$ be the homogeneous representations of $u_i(t_1), u_i(t_2) \in \mathbb{R}^2$. As standard, one can think of these vectors as the vectors from the optical center of the camera to the projection point on the image plane. Using an initial-guess of the pose of the camera at $t_1$, the line passing through $p_E(t_1)$ and ${}^{C_i}q(t_1)$ can be intersected with the DTM. Any ray-tracing style algorithm can be used for this purpose. The location of this intersection is denoted as $^{W}Q_{E_i}$. The subscript letter "$E$" highlights the fact that this ground-point is the estimated location for the feature point, that in general will be different from the true ground-feature location $^{W}Q_i$ (Fig. 2).

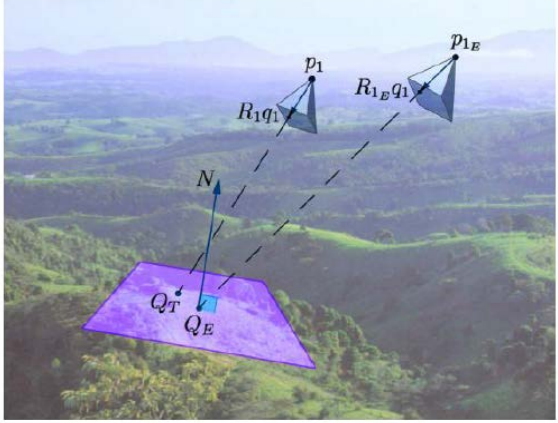

Fig. 2 Ground-feature location

Denoted by $N_i$ the normal of the plane tangent to the DTM at the point $^{W}Q_{E_i}$, one can write:

$$f_i(p_1, \varphi_1, \theta_1, \psi_1, p_{12}, \varphi_{12}, \theta_{12}, \psi_{12}) =$$
$$\left(I - \frac{q_{2i}q_{2i}^T}{q_{2i}^T q_{2i}}\right)\frac{p_{12} + R_{12}\Lambda_i(Q_{E_i} - p_1)}{|p_{12} + R_{12}\Lambda_i(Q_{E_i} - p_1)|} = 0$$
(3)

where $\Lambda_i$ in the above expression represents:

$$\Lambda_i = \frac{q_{1i}N_i^T}{N_i^T R_1 q_{1i}}$$
(4)

This constraint involves the position, orientation and the ego-motion defining the two frames of the camera. For each of the $n$ optical-flow vectors, the function $f_i : \mathbb{R}^{12} \to \mathbb{R}^3$ is defined as the left-hand side of the constraint described in (7). In the above expression, $R_{12}$ and $R_1$ are functions of Euler angles $(\phi_{12}, \theta_{12}, \psi_{12})$ and $(\phi_1, \theta_1, \psi_1)$ respectively. Additionally, the function $F : \mathbb{R}^{12} \to \mathbb{R}^{3n}$ will be defined as the concatenation of the $f_i$ functions:

$$F(p_1, \phi_1, \theta_1, \psi_1, p_{12}, \phi_{12}, \theta_{12}, \psi_{12}) = [f_1, \ldots, f_n]^T.$$

According to these notations, the goal of the algorithm is to find the twelve parameters that minimize $M(\theta, D) = \|F(\theta, D)\|^2$, where $\theta$ represents the 12-vector of the parameters to be estimated and $D$ is the concatenation of all the data obtained from the optical-flow and the DTM. If $D$ were error-free, the true parameters would have been obtained. Since $D$ contains some error perturbation, the estimated parameters are drifted to erroneous values. An iterative scheme will be used in order to solve this system. A robust algorithm which uses Gauss-Newton iterations and M-estimator is described in [20].We begin to use Levenberg-Marquardt method if Gauss-Newton method after several iterations steps to converge. The applicability, accuracy, and robustness of the algorithm were verified through simulations and lab-experiments.

## 3. THE FLIGHT TESTING OF THE PROGRAMME

It was decided to carry out the flight testing of the vision-based navigation programme on the terrain relief with the help of Google Earth video data (based on the real data).

We chose the following conditions of the flight, parameters of the trajectory, the video shooting, and the inertial navigation:

1) The trajectory parameters: The flight is over Xiaoshan district between 30.0421571732 - 30.0545167923 degrees latitude and 120.1245546341- 120.1479005814 degrees longitude. The picture of the district of the flight and the digital map of the flight district are presented in Fig 3. The flight was at a speed of 50 m/sec during 19,6 sec in a straight line at an altitude of 1000 m above mean sea level from the pint (30.05048444, 120.1321713) to the point (30.04588944, 120.1408507). This way is depicted by the red line from the point 1 to the point 50.

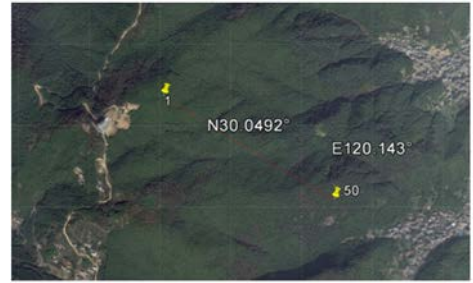

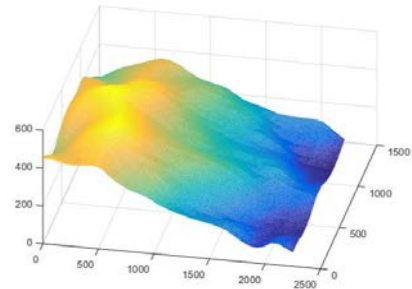

Fig. 3 The district of the flight and the digital map of the flight district.

2) The parameters of the video shooting are as follows:
The 50 images were shot with the 0,4 sec interval between them. The field of view angle of the camera along the short side of the shot is 38,68 degrees; along the long side of the shot - 59,97 degrees. The pixel size of the picture is 4800x2923. The shots are made with the help of the Google Earth. The long side of the shot is perpendicular to the direction of the movement and the airframe, the short side is parallel to the movement, the aircraft flies to the upper side of the picture.

3) The parameters of the inertial navigation system with a random noise are following: the noise results in the average velocity change for the $\Delta V = \pm 20$ m/sec during 1 sec along every axis and average Euler angle change for $\Delta \varphi = \pm 0.33$ degrees during 1 sec. Modeling of the inertial system outputs was carried out with the help of the standard programmes of Matlab INS Toolbox, produced by GPSoft [23].

For the vision-based navigation measuring the index, shots were chosen with time interval about 3.6 sec and an approximate distance from the previous shot of the pair 175m. The characteristic points (about 300) were presented as a regular grid on the shot, the displacement of these points was measured with the help of Lucas and Kanade method with the use of the series of shots between the pair of the index shots for the vision-based navigation (Fig. 4). The arrows show the found displacement of the characteristic points between the pair of shots, used for the vision-based navigation.

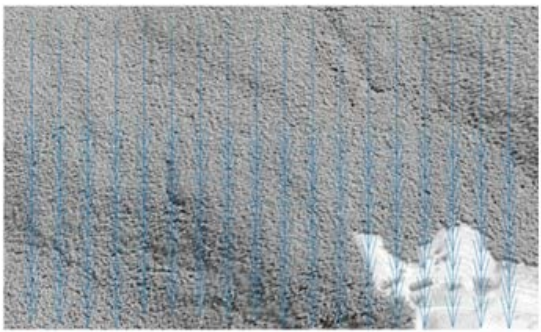

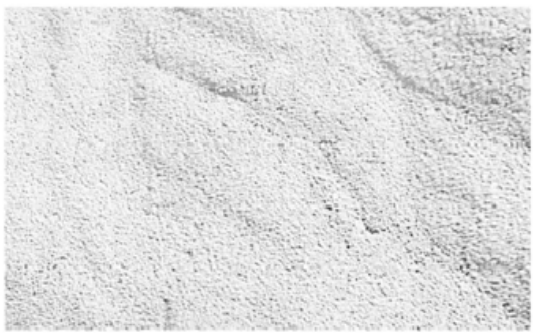

Fig. 4 The pair of shots, used for the vision-based navigation

## 4. GRAPHIC RESULTS OF THE VISION-BASED NAVIGATION PROGRAMME TESTING

On Fig. 5 we present three trajectories: the precise trajectory; the trajectory based on the inertial navigation; the trajectory based on the vision-based navigation.

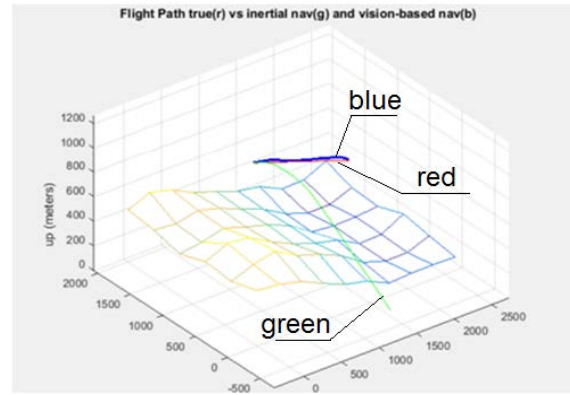

Fig. 5 Three flight trajectories: the precise trajectory is presented in red, the trajectory based on the inertial navigation is presented in green; the trajectory based on the vision-based navigation is presented in blue

The navigation errors are drawn in Fig. 6,7.

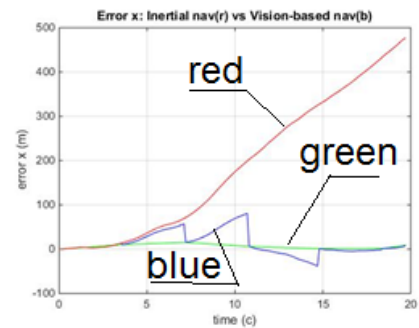

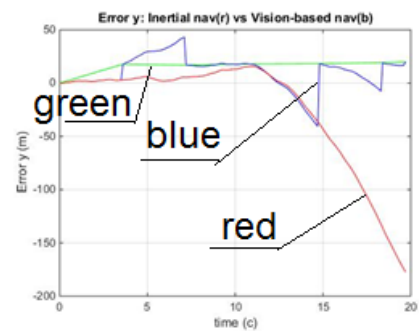

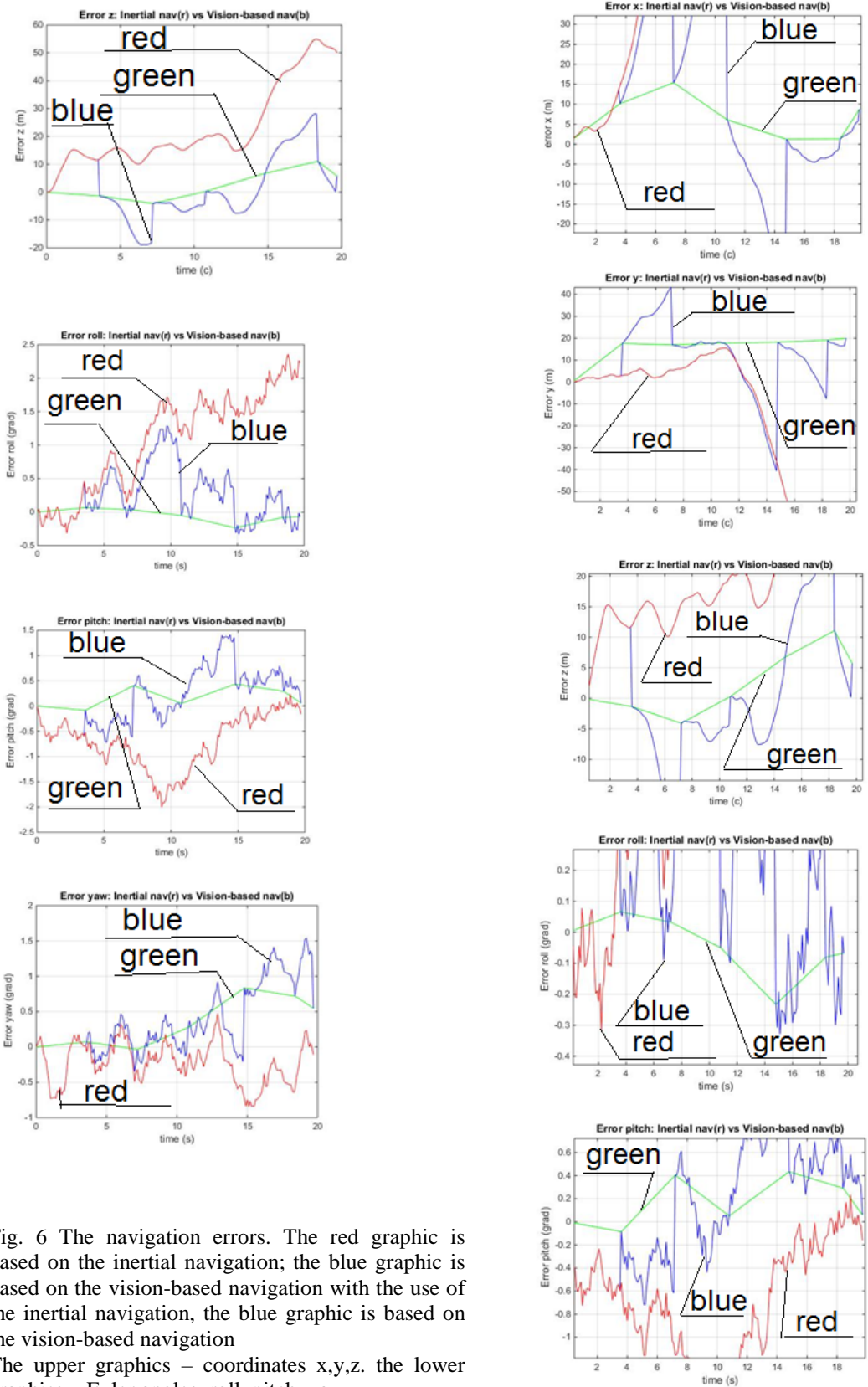

Fig. 6 The navigation errors. The red graphic is based on the inertial navigation; the blue graphic is based on the vision-based navigation with the use of the inertial navigation, the blue graphic is based on the vision-based navigation
The upper graphics – coordinates x,y,z. the lower graphics – Euler angles, roll, pitch, yaw

14

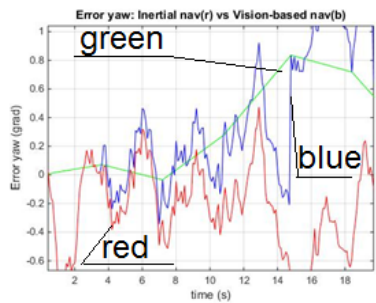

Fig. 7 The same graphic as on Fig 6, but in smaller scale

## 5. CONCLUSION

From the results and discussion given above, the following conclusions can be drawn:

1) The methods of vision-based navigation were used for shots, obtained from an independent source – the programme Google Earth (they are obtained from the flight testing). That means that these results of the vision-based navigation are valid because they are obtained not on the basis of synthetical shots of computer simulation but from an independent source on basis of flight testing.

2) The accuracy of the vision-based navigation system corresponds to the expected for these conditions:

а) Maximum position error based on vision-based navigation is 20 m б) Maximum angle Euler error based on vision-based navigation is 0.83 degree.

3) Also, the measurement without INS usage was carried out:

For all the pairs of shots (used for vision-based navigation) every other shot of the previous pair is the first for the next pair except the first and the last shot.

The camera movement was considered between the two measurements with the help of vision-based navigation method as a straightforward one (without rotation) at a speed found either from positions of the cameras (with the help of the vision-based navigation) for two pairs of shots or from the initial conditions. For this case, the following results were obtained:

а) Maximum position error based on vision-based navigation is 30 m

б) Maximum Euler angle error based on vision-based navigation is 2.2 degrees

The results are worse than with the use of INS.

## 6. ACKNOWLEDGEMENTS